\documentclass[review]{IEEEtran}
\usepackage{amsfonts}
\usepackage{amsmath}
\usepackage{times,amsmath,epsfig,amsmath}
\usepackage{amsfonts,amssymb,amsthm}
\usepackage{epsfig}
\usepackage{float}
\usepackage{graphicx}
\usepackage{color,hyperref}
\usepackage{multirow}
\usepackage{tipa}
\usepackage{cite}
\usepackage{picins}
\usepackage{booktabs}
\begin{document}

\title{DADA: Driver Attention Prediction in Driving Accident Scenarios}

\author{Jianwu Fang, \emph{Member, IEEE}, Dingxin Yan, Jiahuan Qiao, Jianru Xue, \emph{Member, IEEE},  and Hongkai Yu% <-this % stops a space
%\thanks{*This work was supported by the Natural Science Foundation of China (No. 61751308, 61603057, and 61773311).}% <-this % stops a space
\thanks{

$^{1}$J. Fang, D. Yan, and J. Qiao are with the College of Transportation Engineering, Chang'an University, Xi'an, China; J. Fang is also with the Institute of Artificial Intelligence and Robotics, Xi'an Jiaotong University, Xi'an, China
        {(fangjianwu@chd.edu.cn)}.}%
\thanks{$^{2}$J. Xue is with the Institute of Artificial Intelligence and Robotics, Xi'an Jiaotong University, Xi'an, China
        {(jrxue@mail.xjtu.edu.cn)}.}%
        \thanks{$^{3}$H. Yu is with the Department of Electrical Engineering and Computer Science, Cleveland State University, Cleveland, USA
        {(h.yu19@csuohio.edu)}.}%
}

\maketitle
\begin{abstract}
Driver attention prediction is becoming an essential research problem in human-like driving systems. This work makes an attempt to predict the \underline{d}river \underline{a}ttention in \underline{d}riving \underline{a}ccident scenarios  (DADA). However, challenges tread on the heels of that because of the dynamic traffic scene, intricate and imbalanced accident categories. In this work, we design a semantic context induced attentive fusion network (SCAFNet). We first segment the RGB video frames into the images with different semantic regions (i.e., semantic images), where each region denotes one kind of semantic categories of the scene (e.g., road, trees, etc.), and learn the spatio-temporal features of RGB frames and semantic images in two parallel paths simultaneously. Then, the learned features are fused by an attentive fusion network to find the semantic-induced scene variation in driver attention prediction. The contributions are three folds. 1) With the semantic images, we introduce their semantic context features and verified the manifest promotion effect for helping the driver attention prediction, where the semantic context features are modeled by a graph convolution network (GCN) on semantic images; 2) We fuse the semantic context features of semantic images and the features of RGB frames in an attentive strategy, and the fused details are transferred over frames by a convolutional LSTM module to obtain the attention map of each video frame with the consideration of historical scene variation in driving situations; 3) The superiority of the proposed method is evaluated on our previously collected dataset (named as DADA-2000) and two other challenging datasets with state-of-the-art methods.

\end{abstract}


\markboth{Journal of Latex}%
{}

\definecolor{limegreen}{rgb}{0.2, 0.8, 0.2}
\definecolor{forestgreen}{rgb}{0.13, 0.55, 0.13}
\definecolor{greenhtml}{rgb}{0.0, 0.5, 0.0}

\section{Introduction}\label{section1}
\begin{figure}
  \centering
  \includegraphics[width=\hsize]{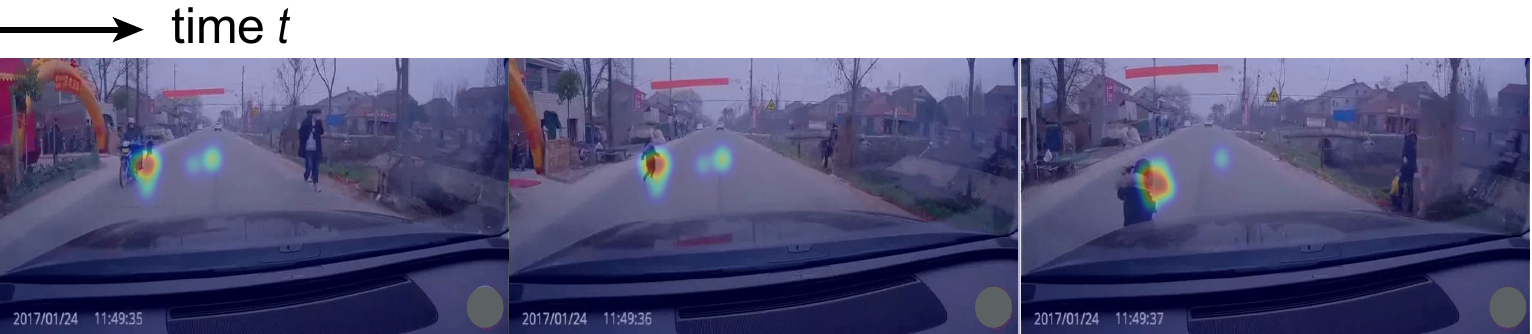}
  \caption{An example of driver attention, where the ego-car hits a crossing girl regretfully. The higher value of the attention map means more allocated attention by drivers.}
  \label{fig1}
  \vspace{-0.3cm}
\end{figure}

\IEEEPARstart{T}{here} are growing research efforts which focus on learning the human sensing mechanism in the vision perception systems of assisted or autonomous driving vehicles, where driver attention is an essential aspect \cite{fridman2018humancentered}. Some previous investigations concluded that one of the main factors of causing road fatalities is the absence of driver attention \cite{gershon2019distracted,cunningham2018driver}, including the distracted driving \cite{Edwards2019distracted}, drowsy driving, drunk driving, etc. Consequently, it is promising to learn the sober human-focusing experience in driving accident scenarios to give a warning of the dangerous object for autonomous driving or assisted driving systems, named as \emph{driver attention prediction in driving accident scenarios} (DADA). Meanwhile, driver attention is one of the important ways to interact with surroundings \cite{rasouli2018joint}, which commonly shows quick identification for the crucial objects or regions (e.g., foveal vision \cite{Perry2002Gaze}) in the crowd traffic scene. Fig. \ref{fig1} shows an example of a person crossing the road, which leads to a collision.

Driver attention has been noticed and studied for decades, and are commonly formulated as searching for the selective road participants in driving situations \cite{DBLP:journals/tits/MorandoVD19,guangyu2019dbus,xia2019periphery}. For a long time, these studies were investigated by a variety of physiological experiments \cite{gaspar2019effect,jha2016analyzing}, such as fatigue detection \cite{DBLP:conf/brain/WanHV13}, but were susceptible to the highly subjective differences between drivers because of the distinct driving habits, driving experience, age, gender, culture, and so on \cite{LedesmaMeasuring,rasouli2018joint}.  Consequently, it is hard to obtain a convincing knowledge to help the driver attention prediction in diverse and different driving scenarios.

Recently, some efforts began to formulate the driver attention prediction as computer vision techniques, and gathered attention data of drivers on large-scale images or videos \cite{palazzi2018predicting,xia2017predicting,DBLP:journals/tits/DengYLY16,deng2019drivers}. For instances, the recent DR(eye)VE project \cite{palazzi2018predicting} collected the driver attention for 555,000 frames in a car (named as in-car collection) mounted eye-tracking equipment. In contrast, the scenarios in DR(eye)VE are sunny and unobstructed, and were exposed by one driver's view in attention collection. In view of this, Berkley Deep Drive laboratory launched a driver attention prediction project (BDDA) in critical situations with braking events \cite{xia2017predicting}. Differently, because of the rather rarity of critical situations, they collected the attention data in a laboratory (named as in-lab collection), and claimed that in-lab collection is better than in-car collection for critical scenarios, because observers are more focused without the disturbance of surroundings and extra maneuvers for controlling cars. BDDA is most related to our work, while it did not consider the driver attention characteristics in actual accident situations in driving.  

In view of this, we constructed a new benchmark with 2000 video sequences (called DADA-2000, with 658,476 frames), laboriously annotated the driver attention and accident intervals. Following BDDA, we also carefully annotated the attention data in lab on various driving scenes with diverse weather conditions (sunny, snowy, and rainy), light conditions (daytime and nighttime), and occasions (highway, urban, rural, and tunnel). DADA-2000 was firstly introduced in our previous work \cite{DBLP:conf/itsc/FangYQXWL19} published in ITSC2019, and has been released on the website \footnote{https://github.com/JWFangit/LOTVS-DADA}. In this version, we made more analysis and updated the ground-truth of attention maps which were obtained by the same criterion with other popular benchmarks, such as DHF1K \cite{wang2019revisiting}. Besides, we design a new model for driver attention prediction in driving accident scenarios.

Specifically, we propose a semantic context induced attentive fusion network (SCAFNet) for driver attention prediction, which is formulated as learning the scene variation that influences the attentional allocation of drivers. To be specific, we first segment the RGB video frames into semantic images \footnote{Semantic image here is obtained by segmenting the RGB video frame into an image with different part of regions, where the pixels in each region denote the same semantic category of scene, such as tree, building, vehicle, pedestrian, etc. Commonly, different semantic regions in one image have distinct proportion.} and learn the spatial-temporal features of the RGB frames and the semantic images by a multi-path 3D convolution network (Sec. \ref{M3DE}) in two parallel paths simultaneously, which is then fused by an attentive fusion network (Sec. \ref{attentive}) to identify the key objects or regions in the scene that attract the drivers' attention potentially. Particularly, the semantic context features of semantic images are learned by a graph convolution network (GCN) to consider the relationship between different semantic parts in the semantic images, and the promotion role of which is verified in this work for the driver attention prediction. We name this component as \emph{semantic context learning of driving scene}. In the attentive fusion network, we consider the transition of features in successive frames by a convolutional LSTM module, so as to consider the historical variation of driving scene when predicting driver attention maps.

We exhaustively evaluate SCAFNet with different components and compare it with state-of-the-art approaches on both our DADA, and two other challenging datasets, i.e., DR(eye)VE  \cite{palazzi2018predicting} and TrafficGaze \cite{deng2019drivers}. In summary, the \textbf{contributions} of this work are three-folds.
\begin{itemize}
\item We propose a new method, i.e., SCAFNet, for driver attention prediction, where the semantic context feature of the driving scene is introduced to help the finding of the key objects/regions that attract drivers' attention, and the semantic context feature is learned by a graph convolution network (GCN) on the semantic images.  
\item We fuse the semantic context features of semantic images and the features of RGB frames in an attentive strategy, and the fused details are transferred over frames by a convolutional LSTM module to identify key objects/regions that attract drivers' attention with the consideration of historical scene variation in driving accident situations.
\item The superiority of the proposed method is demonstrated against some state-of-the-art approaches on three datasets, i.e., our DADA, DR(eye)VE  \cite{palazzi2018predicting}, and  a new dataset TrafficGaze \cite{deng2019drivers}.  As far as we know, this is the first work which focuses on the driver attention prediction in driving accident scenarios.
\end{itemize}

The rest of this paper is organized as follows. Section \ref{relatedwork} briefly reviews the related literatures to this work. Section \ref{dada2000} presents more analysis on DADA-2000, and Section \ref{MSAFNet}  presents the proposed approach. Section \ref{experiments} provides the extensive experiments and analysis, and the final conclusions are given in Section \ref{con}.

%-------------------------------------------------------------------------
\section{Related Work}
\label{relatedwork}

This work is closely related to the visual attention prediction in general images or videos captured from various scenarios and the driver attention prediction in driving scenarios, which are discussed in the following subsections.
\subsection{Visual Attention Prediction}
Visual attention prediction aims to quantitatively localize the most attractive regions in images or video frames by human eyes, commonly producing a 2D saliency map that allocates the likelihood on the locations attracting the fixations \cite{Koch1985Shifts,itti2004automatic,kruthiventi2017deepfix,jiang2018deepvs}, where the images or videos are captured from anywhere without the constraint of scenarios. By that, it is testable to understand the human eye-gazing pattern at the neural activity, e.g., the human interest and the intention. Generally, previous human fixation prediction works can be categorized as top-down methods and bottom-up approaches. Top-down formulations often find the image regions of interest that attract the human attention \cite{borji2018saliency}, which often entails supervised image/video analysis with pre-collected attention data. Bottom-up mechanism commonly detects the salient regions of images or videos in a free-viewing mode \cite{DBLP:journals/corr/abs-1904-09146}. Among these two categories, bottom-up models were extensively studied, and they concentrated on the distinctive image region information of interests in free-viewing. The photometrical \cite{DBLP:journals/pami/ChengMHTH15,itti2004automatic} (e.g., color and contrast), geometrical \cite {CongLFCLH19} (symmetry and center-bias), psychophysical \cite{DBLP:journals/tip/LaiWSS20} (e.g., interestingness and objectness), and psychological \cite{DBLP:journals/pami/ZhangS16} (e.g., the principles in Gestalt) clues are commonly investigated.

Different from the visual attention researches for static images, visual attention prediction in videos concentrates more on motion or object correlations in temporal frames \cite{kruthiventi2017deepfix,jiang2018deepvs,DBLP:journals/tip/ChenWPZQ20,DBLP:conf/aaai/WuWZJW20,chen2019improved,DBLP:journals/tip/LaiWSS20}. With the help of the development of deep learning and the large-scale annotated video database for visual attention prediction, such as Hollywood-2 \cite{DBLP:journals/pami/MatheS15}, UCF sports \cite{DBLP:journals/pami/MatheS15}, DIEM \cite{mital2011clustering}, LEDOV \cite{jiang2018deepvs}, DHF1K \cite{wang2019revisiting}, etc., this field is promoted with a large progress. For instance, Jiang \emph{et al.} \cite{jiang2018deepvs} proposed a saliency-structured convolutional long short-term memory (SS-ConvLSTM) model which considered the spatial center-surround bias and the saccade mechanisum of eye movement over frames. Wang \emph{et al.} \cite{wang2019revisiting} designed an attentive CNN-LSTM model to learn the spatial and temporal scene variation in visual attention prediction. The work \cite{lai2019video} designed a multi-stream fusion network that investigated different fusion mechanisms for spatial and temporal information integration. Recent progress exploited the stimuli from RGB videos and focused on the motion clue largely, while rare work considered the semantic information within the scene videos. As for driving scenarios, semantic variation of scene plays significant role for safe driving \cite{rasouli2018joint,DBLP:journals/tits/MorandoVD19}. 
\subsection{Driver Attention Prediction}
Compared with visual attention prediction in general images or videos, driver attention prediction aims to obtain the attention maps of driving scene videos, and the subject must be drivers. Drivers can quickly identify the important visual clues which influence their driving intention in the blurry periphery vision, and then move the foveal vision of their eyes to the important scene regions \cite{xia2019periphery,xia2019driver}. In other words, driver attention is a direct clue to understand the driver's behavior and the intention in different driving situations \cite{DBLP:journals/tits/MorandoVD19,guangyu2019dbus}. Over decades, driver attention has been widely investigated with the human-designated information \cite{wang2017traffic,DBLP:conf/ivs/ZabihiZBB17}, such as traffic signs, pedestrians, road, as well as other kinds of traffic participants. Benefiting from the progress of saliency computation models, driver attention in the actual driving situation has been exploited in various applications, such as novelty detection \cite{abati2019latent} in driving situations, important object detection in driving \cite{DBLP:journals/tits/WangHXYL17,DBLP:conf/itsc/SchwehrW17}, periphery-fovea driving model designing \cite{xia2019periphery}, and so on.

In order to mimic the driver attention mechanism for large-scale and diverse traffic scenarios, Palazzi \emph{et al.} launched the DR(eye)VE project \cite{palazzi2018predicting} that exploited the driver fixation pattern in an actual car which was exposed to sunny and unobstructed traffic scene, and on this basis, several models based on deep neural networks (e.g., fully connected network (FCN), multi-branch 3D CNN) \cite{deng2019drivers,DBLP:journals/tits/DengYLY16,DBLP:conf/ivs/PalazziSCAC17} were built for driver attention prediction. However, the video frames in DR(eye)VE only collected one driver's gazing behavior which has large subjective difference. Beside DR(eye)VE, there were also some other attempts \cite{DBLP:journals/tits/DengYLY16}, whereas the datasets in these attempts were annotated coarsely and cannot reflect the driver attention convincingly. More recently, Berkeley DeepDrive Laboratory constructed a large-scale driver attention dataset in lab focusing on the critical situations, named as BDDA \cite{xia2017predicting}, and built a simple convolutional neural networks (CNN) to predict the driver attention. BDDA is the most related one to our work, whereas it did not consider the attention process from the critical situations to actual accidents. In our case, we propose a new driver attention prediction model in driving accident scenarios.

%-------------------------------------------------------------------------
\section{DADA-2000 Dataset}
\label{dada2000}

DADA-2000 obtained $658,476$ available frames contained in $2000$ videos with the resolution of $1584\times660$ (about 6.1 hours with $30$ fps, over than DR(eye)VE \cite{palazzi2018predicting}). It is the first dataset that concentrates on the driver attention prediction in driving accident scenarios. The accident occasion in each sequence is a free presentation without any trimming work. In this way, the attention collection may be more natural. The statistics of accident category, scene diversity, sequence numbers of DADA-2000 can be seen in our previous work \cite{DBLP:conf/itsc/FangYQXWL19}. In order to provide a self-sufficient understanding for this work, we present more analysis on the temporal accident window and the attention type of DADA-2000. 
\begin{figure}[htpb]
\centering
\includegraphics[width=0.85\linewidth]{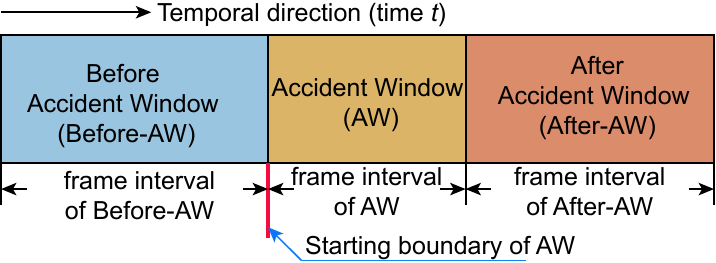}
    \caption{\small{Illustration of the temporal window for each sequence.}}
\label{fig2}
\vspace{-0.1cm}
\end{figure}
\begin{figure}[htpb]
\centering
\includegraphics[width=\linewidth]{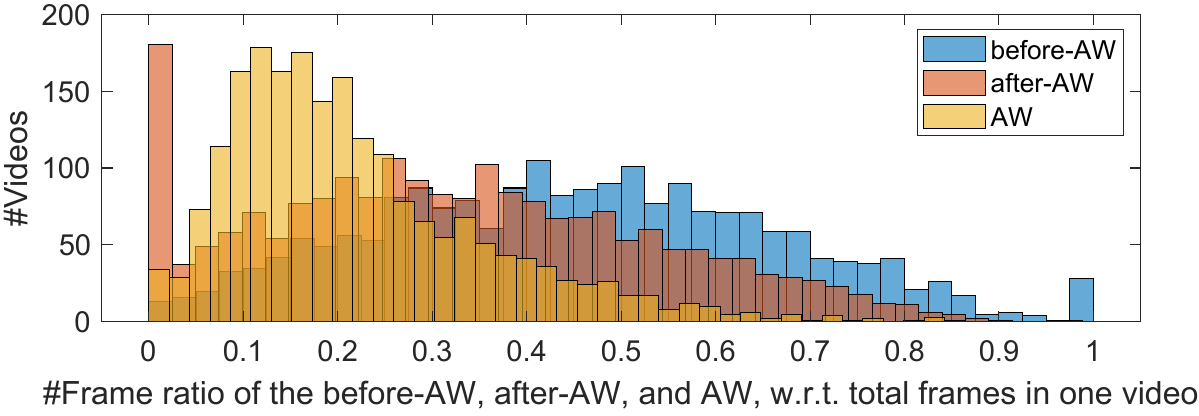}
    \caption{\small{The frame interval distributions of before-AW, after-AW, and AW of DADA-2000. }}
\label{fig3}
  \vspace{-0.3cm}
\end{figure}
\begin{figure}[htpb]
\centering
\includegraphics[width=\linewidth]{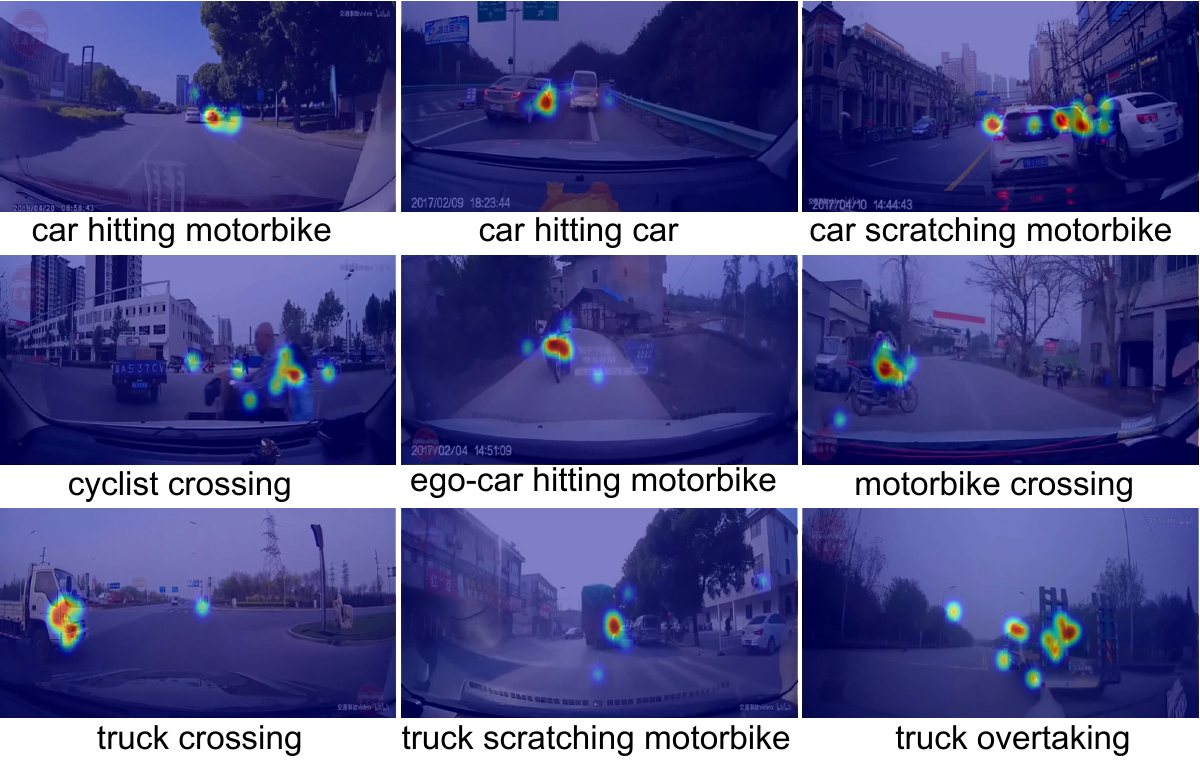}
    \caption{\small{Frame examples with ground-truth attention maps for $9$ kinds of driving accident scenarios.}}
\label{fig4}
  \vspace{-0.3cm}
\end{figure}
\begin{figure*}[!t]
\centering
\includegraphics[width=\linewidth]{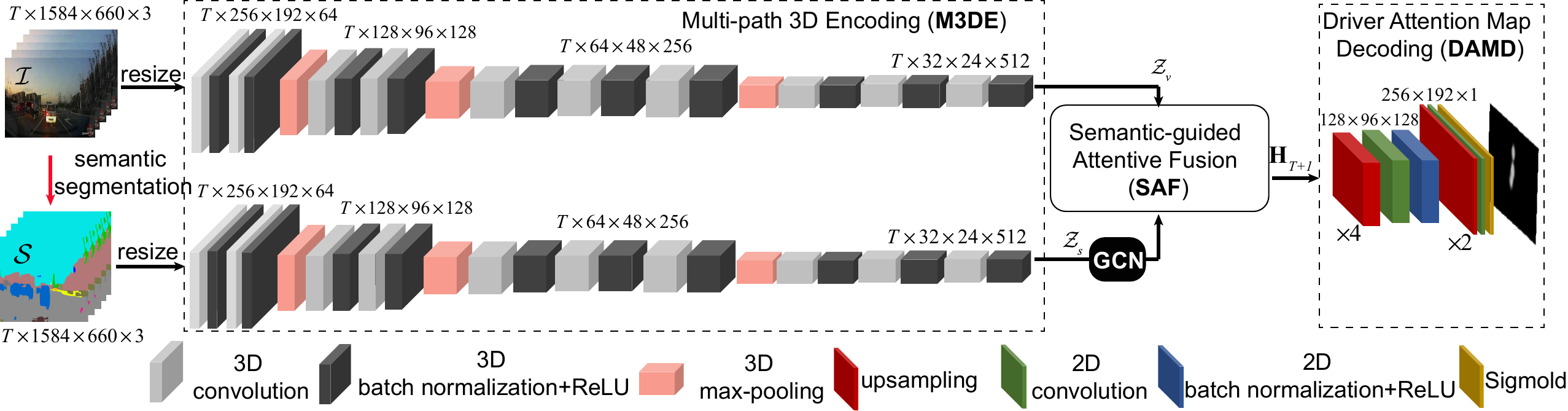}
    \caption{\small{The architecture of the SCAFNet. Given a video clip $\mathcal{I}$ with $T$ RGB frames, we firstly obtain the semantic images $\mathcal{S}$ by popular semantic segmentation models. Then the vision clip and the semantic clip are fed into the multi-path 3D feature encoding pipeline interleaved many kinds of 3D blocks, interleaved in \emph{3D convolution}, \emph{3D batch normalization}+ \emph{rectified linear unit (ReLU)},  and \emph{3D max-pooling} blocks, which generates the features of RGB frame clip ($\mathcal{Z}_{v}$) and semantic image clip ($\mathcal{Z}_{s}$), respectively. Furthermore, $\mathcal{Z}_{v}$ and $\mathcal{Z}_{s}$ are taken into the input of the semantic-guided attentive fusion module which generates the hidden driver attention map ${\bf{H}}_{T+1}$ for the $(T+1)^{th}$ frame. ${\bf{H}}_{T+1}$ is then decoded for the final attention map of the $(T+1)^{th}$ frame with interleaved \emph{upsampling}, \emph{2D convolution}, and \emph{batch normalization+ReLU} operations. It is worthy noting that in this work, we introduce a graph convolution network (GCN) to model the semantic context feature within $\mathcal{Z}_{s}$. The blocks with the same shape have the same size, and the numerical values of each block denote the block height, the block width and the block channels, respectively. (This figure should be viewed in color mode.)}}
\label{fig6}
  \vspace{-0.3cm}
\end{figure*}
\subsection{Temporal Statistics} In DADA-2000, we collected the human fixations for each video frame. For a video sequence, we partitioned it into three main clips: the frame interval before the accident window (before-AW), accident window (AW) and the frame interval after the AW (after-AW), as illustrated in Fig. \ref{fig2}. For AW determining, following the criterion of the work \cite{DBLP:conf/iros/YaoXWCA19}, if about half part of the object (we define it as \textbf{crash-object} in this paper) that will occur accident appears in the camera view, we manually set the frame as the starting boundary of AW, and set the frame as the ending boundary if the scene returns to a normal moving condition or fully stop. The frame interval distributions of before-AW, AW, and after-AW are presented in Fig. \ref{fig3}, where the frame ratio of before-AW, AW, and after-AW in each sequence is computed by the ratio between the number of frames within the specific AW and the total number of frames in the sequence. From these statistics, we find that the frames in accident window are rather fewer than the ones in normal driving.

\subsection{Attention Type}
\label{attention}
In our attention collection protocol, we employed $20$ volunteers with at least $3$ years of driving experience. The eye-tracking movement data were recorded in a laboratory by a Senso Motoric Instruments (SMI) RED250 desktop-mounted infrared eye tracker with $250$ Hz. As described in our previous version \cite{DBLP:conf/itsc/FangYQXWL19}, each video frame was observed by at least $5$ observers for two times. Finally, we recorded about $90$ fixation points for each frame averagely. Following \cite{wang2019revisiting}, we take a Gaussian filter with the size of $50$ pixels to convolute the fixations in each frame into an attention map.  The attention maps for nine kinds of typical driving accident scenarios are shown in Fig. \ref{fig4}.

\subsection{Training/testing Splits}
Because there are over 658,476 frames in DADA-2000, it is very huge and not easy to train. Therefore, in this work, we selected half of the videos (1000 videos) for training, validation and testing. Notably, we still maintain the same number of the accident scenarios (54 categories), even that some ones only have one sequence. Then, we partitioned the selected videos as  the ratio of 3:1:1 for training, validation and testing, i.e., 598 sequences (with about 214k frames), 198 sequences (about 64k frames), and 222 sequences (with about 70k frames), respectively. If some kinds of driving accidents have only one video, we take them into the testing part. In this work, we take the training set and the testing set in the experimental evaluation.

\section{Our Approach}
\label{MSAFNet}
In the driver attention prediction in driving scene, the eye movement of drivers are commonly influenced by the driving tasks and the semantic context  (i.e., the relationship between different semantic scene categories, such as pedestrians, vehicles, road, etc.) variation of scene \cite{rasouli2018joint,DBLP:journals/tits/MorandoVD19}. For example, the driver's attention is easily to be attracted by the sudden movement of the surrounding participants. Therefore, this paper formulates the driver attention prediction model as a semantic context induced attentive fusion network (SCAFNet), which fuses the features of RGB frames and the semantic context features within semantic images to find the key objects or regions that attract the driver's attention. 

Fig.  \ref{fig6} demonstrates the system pipeline of SCAFNet. Assume we have a video clip $\mathcal{I}$ consisting of $T$ RGB frames $\{I_t\}_{t=1}^T$, where $I_t$ is the $t^{th}$ frame. As aforementioned, we first segment the RGB frames within the clip into semantic images $\mathcal{S}=\{S_t\}_{t=1}^T$, and then extract the spatial-temporal features $\mathcal{Z}_{v}=\{{\bf{Z}}^{v}_t\}_{t=1}^T$ (encoded from RGB video frames) and $\mathcal{Z}_{s}=\{{\bf{Z}}^{s}_t\}_{t=1}^T$ (encoded from semantic images) in parallel way by a multi-path 3D encoding (M3DE) architecture (Sec. \ref{M3DE}). Secondly, a semantic-guided attentive fusion module (i.e., SAF in Sec. \ref{attentive}) is modeled to fuse $\mathcal{Z}_{v}$ and $\mathcal{Z}_{s}$, which learns to transfer the fused details over frames and predict the hidden driver attention map ${\bf{H}}_{T+1}$ for the $(T+1)^{th}$ frame. After that,  ${\bf{H}}_{T+1}$ is decoded by a driver attention map decoding module (DAMD) to obtain the final driver attention map. It is worthy noting that ${\bf{H}}_{T+1}$ considers the historical scene variation within previous $T$ frames in the clip. In the following subsections, we will elaborate each module in detail.
\begin{figure*}[!t]
\centering
\includegraphics[width=0.9\linewidth]{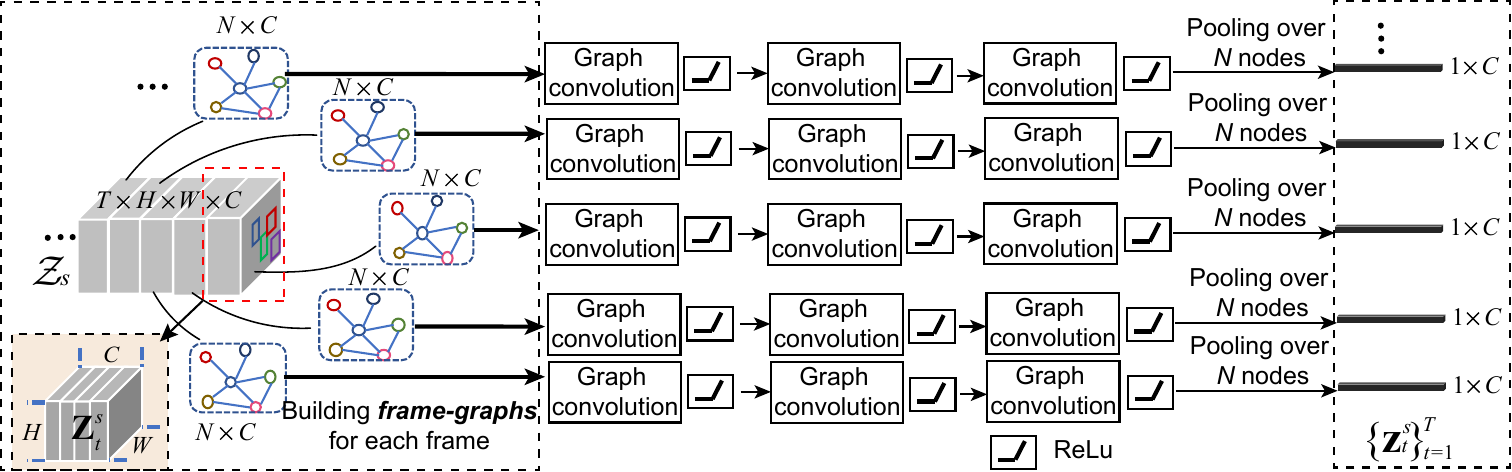}
    \caption{\small{Semantic context learning by graph convolution network (GCN).}}
\label{fig7}
\end{figure*}

\subsection{The M3DE Architecture}
\label{M3DE}

The M3DE architecture aims to extract the spatio-temporal vision and semantic features of driving scene simultaneously. Based on the investigation, semantic clues (e.g., the aforementioned semantic context feature) of the traffic scene is informative for the learning of maneuver behavior for controlling cars, and has been utilized in many recent driving models \cite{li2018rethinking,zhao2019lates}. Considering this, we introduce the semantic information of driving scene to help the driver attention prediction.

For the spatial-temporal feature extraction in this work, we adopt the 3D convolutions to extract the spatial and temporal information over multiple frames because of its easy training property.  As aforementioned, we firstly obtain the semantic images $\mathcal{S}$ of a video clip $\mathcal{I}$. Here, we utilize a popular semantic segmentation approach DeeplabV3 \cite{DBLP:journals/corr/ChenPSA17}  pre-trained by the Cityscapes \cite{cordts2016cityscapes} to fulfill this purpose. Then, as shown in Fig. \ref{fig6}, $\mathcal{S}$ and $\mathcal{I}$ are fed into each path of M3DE, respectively. Each path of M3DE has the same structure, which is constructed by three kinds of interleaved blocks, i.e., \emph{3D convolution} blocks, \emph{3D batch normalization+rectified linear unit (ReLU)} blocks, and  \emph{3D max-pooling} blocks, where 3D batch normalization block is used to accelerate the convergence of network training and resist the gradient vanishing. Totally, each path of M3DE has ten \emph{3D convolution} blocks, ten \emph{3D batch normalization+ReLU} blocks, and three \emph{3D max-pooling} blocks. Notably, Fig. \ref{fig6} demonstrates the size of each block, where the numerical values of each block specify the block height, the block width and the block channels, respectively. 

In our implementation, we resize the consecutive frames of input into the resolution of $256\times192$. The kernel size, stride size, and padding size of the \emph{3D convolution} operation are $3\times3\times3$, $1$, and $1$, respectively. After passing M3DE architecture, we obtain the spatio-temporal features $\mathcal{Z}_{v}=\{{\bf{Z}}^{v}_t\}_{t=1}^T$ of $\mathcal{I}$ and $\mathcal{Z}_{s}=\{{\bf{Z}}^{s}_t\}_{t=1}^T$ of $\mathcal{S}$, and then they are fed into the following SAF module. Notably, $\mathcal{Z}_{s}$, $\mathcal{Z}_{v}$ $\in\mathbb{R}^{T\times 32\times 24\times512}$ are 4D tensors owning 512 feature channels with $32$$\times$$24$ resolution for $T$ frames.

\subsection{SAF Module}
\label{attentive}
With the feature $\mathcal{Z}_{v}$ and $\mathcal{Z}_{s}$ of RGB frames and semantic images, we insightfully design a fusion strategy to explore the complementary characteristics of them. As we all know that different semantic categories (e.g., pedestrians, vehicles, road, etc.) in driving scene have distinct importance to drivers \cite{DBLP:journals/pr/Ohn-BarT17a}, and the importance of semantic categories is commonly determined by their relations in the scene (denotes as \emph{semantic context} in this work). For example, if a person stands in the middle of the road, the person is important to drivers, which implies an irregular relationship priori between the person and the road for normal driving. Therefore, this work introduces the semantic context of driving scene for driver attention prediction. As for the relationship modeling, graph convolution network (GCN) \cite{DBLP:journals/ijcv/VeitB20} is currently a powerful tool to learn the deep relations within the data. Hence, this work models the relationship between different semantic categories of scene by a graph convolution network (GCN) inspired by the work of \cite{DBLP:conf/eccv/WangG18} to extract the semantic context feature of the driving scene. By that, we can find the important relations between semantic categories of the driving scene for driver attention prediction. Nevertheless, different from \cite{DBLP:conf/eccv/WangG18}, we build the graph frame by frame independently. That is because we want to provide a basis for the following fusion module, where each frame has both the semantic context feature and the spatial-temporal feature of RGB frame in fusing process.

In addition, since this work focuses on the driver attention prediction in driving accident scenarios, we want to memorize and transfer the features of potentially key objects or regions in successive $T$ frames to the $(T+1)^{th}$ frame for a prediction. For example, if there is a pedestrian crossing situation, we want to make the attention value on the pedestrian larger and larger with more frame learning. Based on the tensor form of $\mathcal{Z}_{v}$ and $\mathcal{Z}_{s}$, we adopt the convolutional LSTM (Conv-LSTM) \cite{DBLP:conf/nips/ShiCWYWW15} to learn and transfer the fused details within successive $T$ frames to the $(T+1)^{th}$ frame (we call this as \emph{transition of spatio-temporal scene features}). Although Conv-LSTM in our case fulfills a many-to-one prediction for driver attention map, we can generate the attention map for each video frame by densely sampling the $T$ frames sequentially along each video sequence in testing stage. It is similar to the setting of the famous DR(eye)VE \cite{palazzi2018predicting}. 

In the following, we will describe how to learn the semantic context of driving scene and the transition of spatio-temporal scene features. Then, the so-called semantic-guided attentive fusion strategy is presented.

\subsubsection{Semantic Context Learning of Driving Scene}
For the semantic context learning of driving scene, we build a graph over feature of each semantic image (i.e., ${\bf{Z}}^{s}_t$ for the $t^{th}$ frame) independently, named as \emph{frame-graph}. In fact, after the 3D convolution in M3DE architecture, ${\bf{Z}}^{s}_t$ has short-term spatial-temporal information of consecutive semantic images. Fig. \ref{fig7} demonstrates the semantic context learning framework, the main components of which are the \emph{frame-graph} building and graph convolution operation.  

\textbf{Frame-graph building:}
Formally, assume we obtain the semantic features $\mathcal{Z}_{s}=\{{\bf{Z}}^{s}_t\}_{t=1}^T\in \mathbb{R}^{T\times H\times W \times C} $ of $T$ semantic images, where $H$, $W$, and $C$ denote the height, the width and the number of channels of ${\bf{Z}}^{s}_t$, respectively  (shown in Fig. \ref{fig7}).  For the semantic feature ${\bf{Z}}^{s}_t\in \mathbb{R}^{H\times W \times C}$ of the $t^{th}$ frame, we re-formulate it as a matrix ${\bf{\hat{Z}}}^{s}_t\in \mathbb{R}^{N \times C}$, where $N=H\times W$ represents the number of the nodes in the frame-graph. In order to build the graph, we need to measure the pair-wise node similarity within ${\bf{\hat{Z}}}^{s}_t$, defined as: 
\begin{equation}
{\bf{S}}_t^s=\phi({\bf{\hat{Z}}}^{s}_t)^T\phi^{'}({\bf{\hat{Z}}}^{s}_t),
\end{equation}
where $\phi({\bf{\hat{Z}}}^{s}_t) ={\bf{W}}_T {{\bf{\hat{Z}}}^{s}_t} $ and $\phi^{'}({\bf{\hat{Z}}}^{s}_t) ={\bf{W}}_T^{'} {{\bf{\hat{Z}}}^{s}_t} $ represent the linear transformation of ${\bf{\hat{Z}}}^{s}_t$, which is fulfilled by a fully connected layer in our case. ${\bf{S}}_t^s\in  \mathbb{R}^{N\times N}$ is the similarity matrix of nodes. ${\bf{W}}_T$, ${\bf{W}}_T^{'}\in \mathbb{R}^{C\times C}$ are the weight matrixes of the transformations $\phi$ and $\phi^{'}$, respectively. Motivated by the works \cite{DBLP:conf/eccv/WangG18,DBLP:conf/nips/VaswaniSPUJGKP17}, we normalize ${\bf{S}}_t^s$ and obtain the affinity matrix ${\bf{G}}^s_t\in \mathbb{R}^{N\times N}$ by the \emph{softmax} function in the column dimension of ${\bf{S}}_t^s$. So far, the affinity matrix ${\bf{G}}^s_t$ is constructed as the \emph{frame-graph}. 

\textbf{Graph convolutions on frame-graph:}
Graph convolution computes the relations of the nodes with their neighbors. In our case, our GCN has three layers of graph convolution to explore the semantic relation over different nodes in the graph ${\bf{G}}^{s}_t$ for the $t^{th}$ frame. The graph convolution operation is defined as:
\begin{equation}
{\bf{\bar{Z}}}^{s}_t={\bf{G}}^{s}_t {\bf{\hat{Z}}}^{s}_t {\bf{W}}^{s}_t,
\label{eq:3}
\end{equation}
where ${\bf{W}}^{s}_t \in \mathbb{R}^{C\times C}$ denotes the weight of the graph convolution layer, and ${\bf{\bar{Z}}}^{s}_t \in \mathbb{R}^{N\times C}$ is the output of each graph convolution layer. In this work, the output ${\bf{\bar{Z}}}^{s}_t$ is fed into next layer after a \emph{Relu}. In the final layer of graph convolution, the output is followed by an averaging operator over $N$ nodes and generates a vector ${\bf{z}}^{s}_t \in \mathbb{R}^{1\times C}$, as illustrated in Fig. \ref{fig7}. Notably, ${\bf{z}}^{s}_t$ is a sparse vector commonly, and the non-zero elements encode the preferred semantic context information.  Consequently, for $\mathcal{Z}_{s}=\{{\bf{Z}}^{s}_t\}_{t=1}^T$, we conduct the GCN on each ${\bf{Z}}^{s}_t$, and obtain the semantic context features $\{{\bf{z}}^{s}_t\}_{t=1}^T$ for $T$ frames, As defined in M3DE architecture, $H, W, C$ are $32$, $24$, $512$, respectively.
\subsubsection{Transition of Spatio-temporal Scene Features}
Conv-LSTM extends the LSTM by preserving the spatial details in predicting when learning the temporal features. It uses the memory cell ${\bf{C}}_t$ and the hidden state cell ${\bf{H}}_t$ of time $t$ (frame index in our case) to control the memory updating and the output of ${\bf{H}}_t$, and transfers the spatio-temporal scene features sequentially (Here, we omit the distinction for vision path and semantic path). Precisely, a ConvLSTM module is defined as:
\begin{equation}
[{\bf{C}}_{t+1},{\bf{H}}_{t+1}]=\textit{ConvLSTM}({\bf{Z}}_{1:{t}},{\bf{W}}_c),
\label{eq:4}
\end{equation}
where ${\bf{Z}}_{1:t}$ is the input sequence up to time $t$, ${\bf{W}}_c$ is the weight parameter to be learned. The structure of ConvLSTM is defined as the following recursive formulations:
\begin{equation}
\begin{array}{ll} 
            {\bf{i}}_t = \sigma(w_{zi}*{\bf{Z}}_t + b_{zi} + w_{hi} *{\bf{H}}_{t} + b_{hi}) \\
            {\bf{f}}_t = \sigma(w_{zf}* {\bf{Z}}_t + b_{zf} + w_{hf} *{\bf{H}}_{t} + b_{hf}) \\
            {\bf{g}}_t = \tanh(w_{zg} *{\bf{Z}}_t + b_{zg} + w_{hg} *{\bf{H}}_{t} + b_{hg}) \\
            {\bf{o}}_t = \sigma(w_{zo} *{\bf{Z}}_t + b_{zo} + w_{ho} *{\bf{H}}_{t} + b_{ho}) \\
            {\bf{C}}_{t+1} = {\bf{f}}_t\circ {\bf{C}}_{t} + {\bf{i}}_t \circ {\bf{g}}_t \\
            {\bf{H}}_{t+1} = {\bf{o}}_t\circ \tanh({\bf{C}}_{t+1}), \\
        \end{array}
\end{equation}
where ${\bf{Z}}_t$ is the input in the $t^{th}$ frame, and $\sigma$, $tanh$ are the activation functions of logistic sigmoid and hyperbolic tangent, respectively. $``*"$ and $``\circ"$ denote the convolution operator and Hadamard product, and $ {\bf{i}}_t , {\bf{f}}_t , {\bf{o}}_t $ are the convolution gates controlling the input, forget and output, respectively. $[w_{zi},w_{hi},w_{zf},w_{hf},w_{zg},w_{hg},w_{zo},w_{ho}]\in {\bf{W}}_c$. Notably, ${\bf{H}}_{1}$ and  ${\bf{C}}_{1} $ are initialized as 3D tensors with zero elements owning the same dimension to the input ${\bf{Z}}_t$, i.e., with $32\times24\times 512$ dimensions. Eq. \ref{eq:4} fulfills the transition of spatio-temporal features over $t$ successive frames to the ${(t+1)}^{th}$ frame.
\begin{figure}[!t]
\centering
\includegraphics[width=0.8\linewidth]{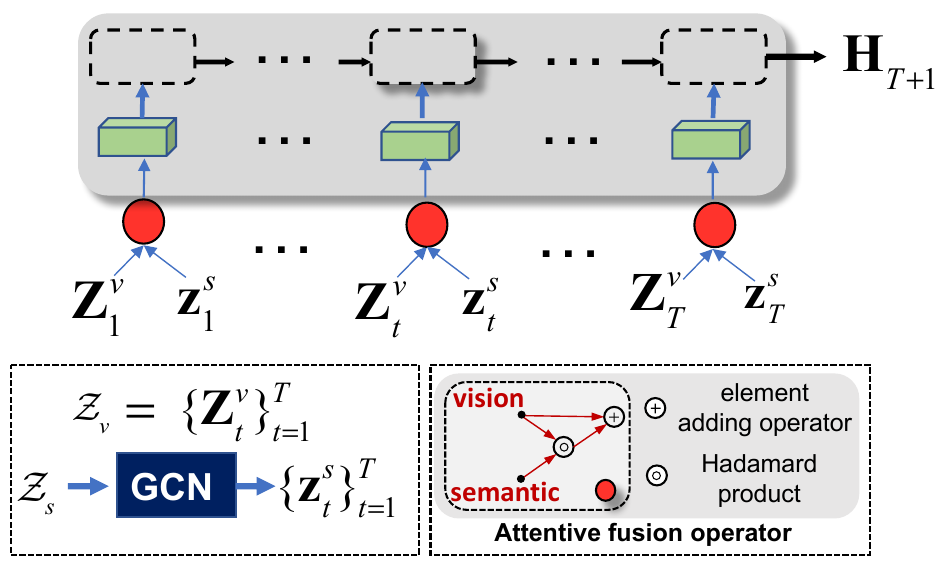}
    \caption{\small{Attentive fusion module.}}
\label{fig8}
\end{figure}
\subsubsection{Attentive Fusion}
\begin{figure}
\centering
\includegraphics[width=\linewidth]{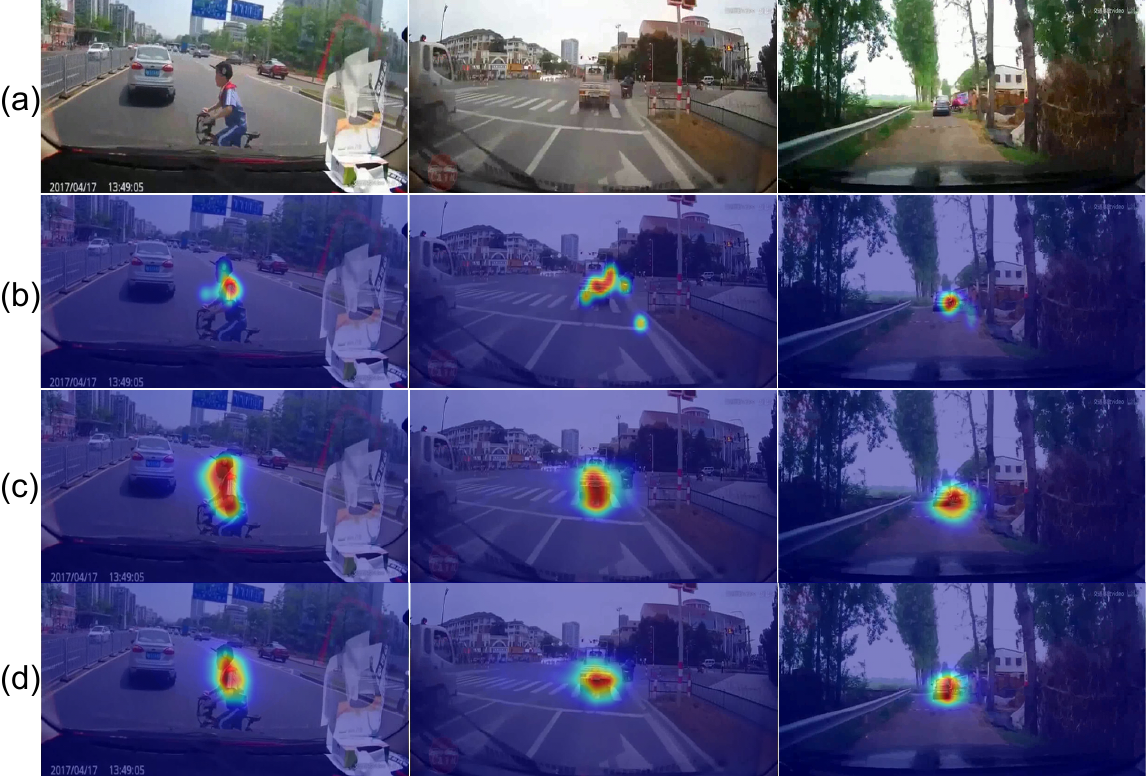}
    \caption{\small{The predicted attention results by (c) the full model without our GCN and (d) the full model with our GCN. (a) and (b) are the original images and the ground-truth attention maps. From this figure, we can see that GCN can make the predicted attention be more focused than the one without GCN. Note that, these results are obtained by running the models with the same configuration.}}
\label{fig9}
\end{figure}

Based on the above modeling, we obtain the features $\{{\bf{Z}}^{v}_t\}_{t=1}^T$ of RGB video frames and the semantic context features $\{{\bf{z}}^{s}_t\}_{t=1}^T$ of semantic images. Then, we will fuse them together to fulfill the semantic-guided representation of driving scene. Inspired by the recent attention mechanism \cite{wang2019revisiting,lai2019video,DBLP:journals/corr/abs-1904-09146}, we introduce a residual connection to maintain the original information in RGB frames after fusion. Therefore, we define the attentive fusion strategy as:
\begin{equation}
[{\bf{C}}_{T+1}, {\bf{H}}_{T+1}]=\textit{ConvLSTM}({\bf{Z}}^{v}_{1:T} \circ (1+{\bf{z}}^{s}_{1:T}), {\bf{W}}_c), 
\end{equation}
where ${\bf{H}}_{T+1}$ is the obtained hidden driver attention map of the $(T+1)^{th}$ frame, and $``\circ"$ denotes the Hadamard product. In our case, we use $256$ hidden states and $3\times3$ convolution kernel for the ConvLSTM layer to generate ${\bf{H}}_{T+1}$ with $32\times24\times 256$ dimensions. Fig. \ref{fig8} demonstrates the attentive fusion module. By this kind of semantic-guided attentive fusion, we fulfill a feature selection which maintains the original information of vision path. Fig. \ref{fig9} demonstrates some results predicted by our method with or without GCN module. From this comparison, we can see that our GCN makes the predicted attention be more focused than that without GCN.
\subsection{DAMD module}
After obtaining the hidden driver attention map ${\bf{H}}_{T+1}$ for the $(T+1)^{th}$ frame, it is fed into the driver attention map decoding module (DAMD) to generate final driver attention map $\hat{Y}$ of the $(T+1)^{th}$ frame. It is worthy noting that the driver attention map of the $(T+1)^{th}$ frame considers the historical scene variation in previous $T$ frames. The structure of DAMD is constructed by six layers, interleaved by \emph{upsampling}, \emph{2D convolution}, and \emph{2D batch normalization} layers.  More specifically, DAMD module is implemented as \emph{upsampling}($\times$4)$\rightarrow$ \emph{convolution} (3$\times$3, 128) $\rightarrow$ \emph{batch normalization+ReLU} $\rightarrow$ \emph{upsampling}($\times$2) $\rightarrow$ \emph{convolution}(3$\times$3, 1) $\rightarrow$ \emph{Sigmold}, where the deconvolution is denoted as convolution (kernel, channels). \emph{Sigmoid} function is used to limit the output value of driver attention map to $[0,1]$.
\subsection{Loss Function}
In this work, we have the human fixations $F$, ground-truth driver attention map $Y$, and the predicted attention map $\hat{Y}$. In this paper, we not only want the predicted $\hat{Y}$ to be approximate to $Y$, but also to make the point with the largest value of $\hat{Y}$ close to $F$.  Formally, our loss function is defined as:
\begin{equation}
\begin{array}{ll} 
\mathcal{L}(Y,\hat{Y})= \sum_{i} Y(i) \log \left(\epsilon+\frac{Y(i)}{\epsilon+\hat{Y}(i)}\right)-\frac{{cov}(Y, \hat Y)}{\rho(Y) \rho(\hat Y)}\\
\quad  \quad\quad\quad -\frac{1}{|P(i)\not=0|}\sum_i \frac{Y(i)-\mu(Y)}{\rho(Y)}F(i),
 \end{array}
\end{equation}
where the first term is the \emph{Kullback-Leibler divergence} (KLdiv) which evaluates the distribution distance of two maps. The second term specifies the \emph{linear correlation coefficient} (CC) measuring the linear relationship between $Y$ and $\hat{Y}$. The last term denotes the \emph{normalized scanpath saliency} (NSS) which prefers a close distance between the points in $Y$ with peak value and the locations of $P$ with non-zero value. $|P(i)\not=0|$ represents the number of points in $P$ with non-zero value. ${cov}(Y, \hat Y)$ is the covariance of $Y$ and $\hat Y$, ${\rho(\cdot)}$ refers to standard deviation, the summation index $i$ spans across image pixels and $\epsilon$ is a small constant that ensures numerical stability. Because CC and NSS prefer a large value for similar maps, they are computed in their negative values.

\section{Experiments}
\label{experiments}
\subsection{Datasets and Implementation Details}
\textbf{Datasets}: In this work, because of the special topic on the driver attention prediction in driving accident scenarios, we firstly evaluate the performance of the proposed method on our DADA dataset, as described in Sec. \ref{dada2000}. Then, in order to verify the superiority of the proposed method, we also evaluate the performance on other two challenging datasets, i.e., the famous DR(eye)VE \cite{palazzi2018predicting} and TrafficGaze \cite{deng2019drivers}.

DR(eye)VE \cite{palazzi2018predicting} contains $555,000$ frames with 74 sequences. It was collected in the practical driving process by eight drivers independently. The eye-movement data of the videos is captured by \emph{Senso Motoric Instruments (SMI) ETG 2w} Eye Tracker with the frequency of 60Hz. The video form in DR(eye)VE is $720p/30fps$.  Similar to the configuration of DR(eye)VE, we use the first half of sequences as the training set and the second half of them as the testing set. 

TrafficGaze \cite{deng2019drivers} consists of $16$ driving videos, where the videos lasted from $52$ to $181$ seconds. TrafficGaze is captured by \emph{Eyelink 2000} (SR Research, Eyelink, Ottawa, 139 Canada) at 1000 Hz, and the frame resolution of which is $1280\times 720$. In TrafficGaze, there are 49,000 frames for training and 19,140 frames for testing.

\textbf{Implementation details}:
During training, we adopted the Adam optimizer with the learning rate of 0.0001 for our DADA and DR(eye)VE datasets.  As for TrafficGaze, the learning rate is set as 0.00001. The whole model is trained in an end-to-end manner, and the entire training procedure takes about 20 hours, 30 hours, and 6 hours on the platform of two NVIDIA RTX2080Ti GPUs and 22GB RAM for DADA, DR(eye)VE and TrafficGaze datasets, respectively. In fact, in order to fulfill the 3D convolution in M3DE module in our method, it needs at least 3 successive frames of input. In addition, based on our previous research \cite{framepre2020},  we have found that for the frame or map prediction problem, much more input frames will take more computation resources while have little performance gain. Therefore, In our implementation, because of the limitation of the RAM space, each video training batch has $2$ clips randomly selected from the training set, and each clip contains $5$ successive frames. 
\begin{table}[!t]\small
\centering
\caption{\small{The formulations of the evaluation metrics. The symbol $\downarrow$ pursues a lower value and $\uparrow$ prefers a higher value. $Y$ and $\hat{Y}$ denote the ground-truth attention map and the predicted one. $F$ is the human fixations of an image. ${cov}(Y, \hat Y)$ is the covariance of $Y$ and $\hat Y$, and ${\rho(\cdot)}$ refers to the standard deviation, where the summation index $i$ spans across image pixels and $\epsilon$ is a small constant that ensures numerical stability. FPR and TPR are the false positive (FP) rate and true positive (TP) rate, respectively. TN and FN represent the number of the true negative pixels and false negative pixels, respectively.}}
\begin{tabular}{c|c|c}
\toprule
Metrics&Formulations&Value Range\\
\hline\hline
KLdiv$\downarrow$& $\sum_{i} Y(i) \log \left(\epsilon+\frac{Y(i)}{\epsilon+\hat{Y}(i)}\right)$&$>$0 \\
\hline
NSS$\uparrow$  &$\frac{1}{|P(i)\not=0|}\sum_i \frac{Y(i)-\mu(Y)}{\rho(Y)}F(i)$ &$>$0 \\
\hline
SIM $\uparrow$ &$\sum_i min(Y(i),\hat{Y}(i))$ &[0,1] \\
\hline
CC$\uparrow$  &$\frac{{cov}(Y, \hat Y)}{\rho(Y) \rho(\hat Y)}$ &[-1,1]  \\
\hline
ROC  &$FPR=\frac{FP}{FP+TN}$, $TPR=\frac{TP}{TP+FN}$ &AUC$\uparrow$$\in$[0,1] \\
\hline
\end{tabular}
\label{tab1}
\end{table}

\subsection{Evaluation Protocols}
\textbf{Evaluation metrics:} Following the existing works \cite{wang2019revisiting,lai2019video,DBLP:journals/corr/abs-1904-09146}, five kinds of metrics are utilized: Kullback-Leibler divergence (KLdiv), normalized scanpath saliency (NSS), similarity metric (SIM), linear correlation coefficient (CC), and the {\bf a}rea  {\bf u}nder the receiver operating characteristic (ROC)  {\bf c}urve (AUC). Here, two variants of AUC,  i.e., AUC-Judd (AUC-J) \cite{DBLP:conf/iccv/JuddEDT09} and shuffled AUC (AUC-S) \cite{DBLP:conf/iccv/BorjiTSI13}, are used. Table. \ref{tab1} gives the mathematical formulations of these metrics. The physical meaning of them are as follows.

KLdiv measures the information loss of the probability distribution of the predicted maps to the ones of the ground-truth, and the smaller value indicates a less information loss.

NSS computes the average value of the positive positions in the predicted attention map, which measures the hitting rank to the ground-truth fixations, and higher value is better. 

SIM concerns about the interaction of the predicted attention map and the ground-truth, which pursues a similar probability distribution, and larger value means a better approximating.

CC calculates the linear relationship of the random variables in two distributions, and similarly the higher value shows a better matching of the distributions.

AUC-J concentrates on the true positive rate (TP rate) and the false positive rate (FP rate), instead of the number of pixels of TP and FP in original AUC. AUC-S aims to overcome the center bias in AUC computation and it compensates for the central fixation bias. These AUCs prefer a larger value for better performance, which are obtained by computing the FP, TP, FN, and TN values between different binary segments of the predicted attention map with the human fixations in distinct binarization thresholds ranging from [0,1].

\begin{figure*}
\centering
\includegraphics[width=\linewidth]{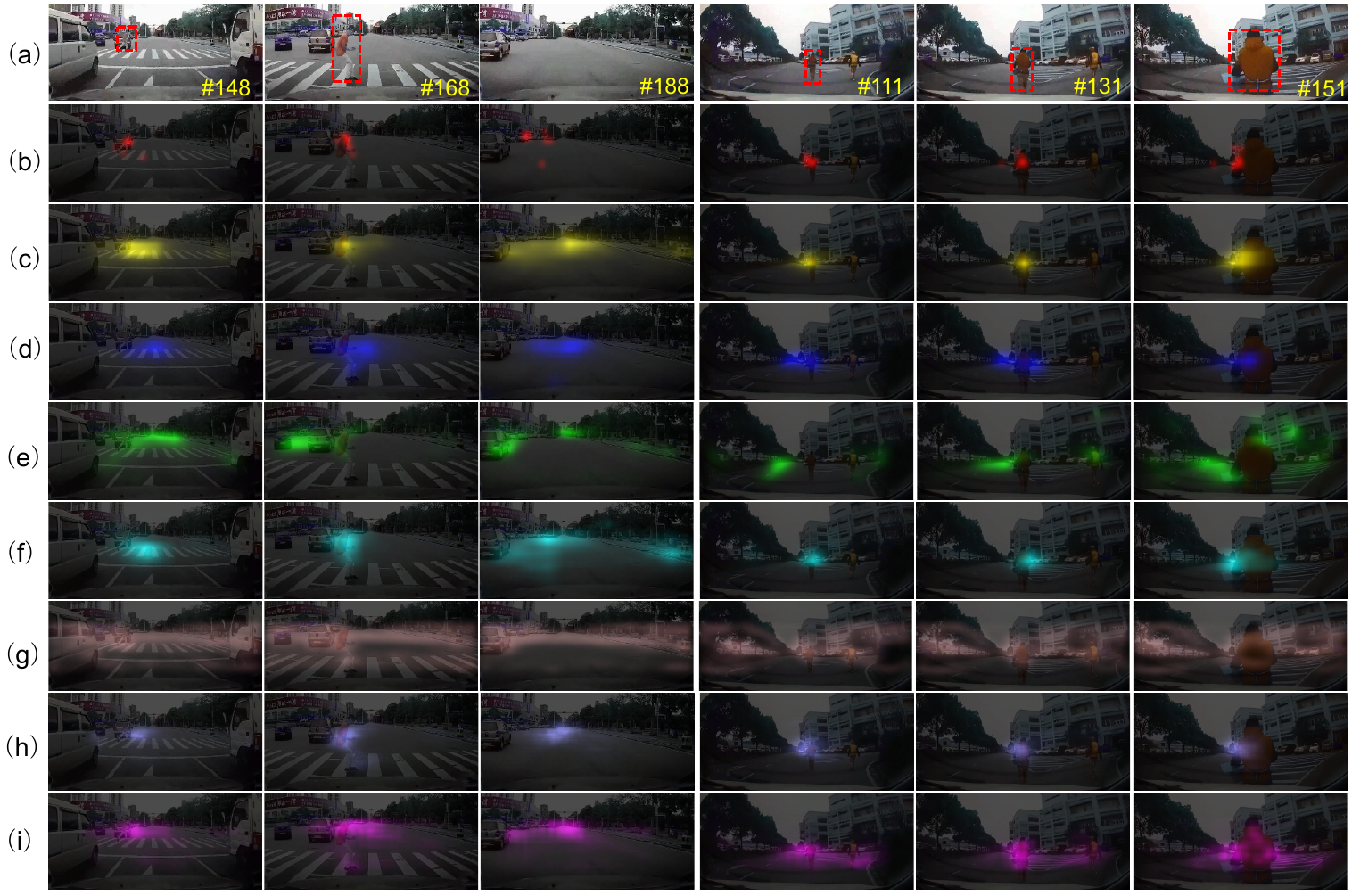}
    \caption{\small{The visualized snapshots demonstrating the attention prediction results by (c) our method, (d) ACLNet \cite{wang2019revisiting}, (e) BDDA \cite{xia2017predicting}, (f) DR(eye)VE \cite{palazzi2018predicting}, (g) TwoStream \cite{bak2017spatio}, (h) SalGAN \cite{pan2017salgan}, and (i) SALICON \cite{huang2015salicon}. (a) and (b) are the original images and the groundtruth. Note that the red bounding boxes label the crash-object. (This figure should be viewed in color mode.)}}
\label{fig10}
  \vspace{-0.2cm}
\end{figure*}

\subsection{Comparison Baselines for DADA Dataset}
In order to analyze the role of different components in the proposed method, we present three different baselines. The proposed SCAFNet is actually a two-branch model which learns the spatial-temporal features of RGB frames and semantic images, simultaneously, and then the features are fused by an attentive fusion module. For the semantic branch, we propose a graph convolution network to model the semantic context of driving scene. Therefore, the first baseline of SCAFNet is the one-branch version that only learns the features of RGB frames without the semantic path, named as \textbf{Ours-One-Branch (Ours-OB.)}. We also want to know the effect of GCN for this work, named as \textbf{Ours-Two-Branch-GCN (TB.-GCN)} and \textbf{Ours-Two-Branch-without-GCN (Ours-TB.-w/o-GCN)}. Actually, Ours-Two-Branch-GCN denotes the full model of SCAFNet.

In addition, in order to verify the superiority of the proposed method, we introduce seven attention prediction methods representing the state-of-the-arts, consisting of five ones for videos, i.e., BDDA \cite{xia2017predicting}, DR(eye)VE \cite{palazzi2018predicting}, TwoStream \cite{bak2017spatio}, MLNet \cite{cornia2016deep}, ACLNet \cite{wang2019revisiting}, and two ones for static images, i.e., SALICON \cite{huang2015salicon} and SalGAN \cite{pan2017salgan}.  Among them, BDDA and DR(eye)VE are two methods that concentrate on the driver attention prediction in critical situations and normal scenarios, respectively, and other ones focus on the general video attention prediction problem. The codes of these approaches are downloaded from their official website and re-trained by our DADA benchmark with the same configuration stated in their works. The original configuration of DR(eye)VE had three branches, i.e., the RGB channel, the semantic channel, and the optical flow channel. In this work, we take the RGB and semantic channel the same as our case to re-train their model.
\subsection{Overall Evaluation on DADA}

In the overall performance evaluation, we compare the baselines on overall testing set of DADA (i.e., $71107$ frames in our case). Table. \ref{tab2} demonstrates the performance of all compared methods on overall testing set of DADA.  Fig. \ref{fig10} demonstrates a qualitative evaluation, where some frame snapshots of different approaches are presented. 
\begin{table}\vspace{0.3cm}\footnotesize
\centering
\caption{\small{Performance comparison on the overall testing set of our DADA dataset.}}
\begin{tabular}{c|c|c|c|c|c|c}
\toprule
Methods/Metrics &Kldiv&NSS &SIM &CC&AUC-J&AUC-S\\
\hline\hline
SalGAN \cite{pan2017salgan}&3.57&3.21&0.37&0.48 &0.91&0.64 \\
SALICON \cite{huang2015salicon} &\textbf{2.17} &2.71&0.30&0.43 &0.91&0.65 \\
TwoStream \cite{bak2017spatio} &2.85 &1.48 &0.14 &0.23&0.84 &0.64 \\
MLNet \cite{cornia2016deep}&11.78 &0.30 &0.07 &0.04 &0.59 &0.54 \\
BDDA \cite{xia2017predicting}&3.32 &2.15 &0.25&0.33 &0.86 &0.63 \\
DR(eye)VE \cite{palazzi2018predicting}&2.27 &2.92 &0.32 &0.45 &0.91 &0.64 \\
ACLNet \cite{wang2019revisiting}&2.51 &3.15 &0.35 &0.48 &0.91 &0.64 \\
\hline
Ours-OB. &2.58 &3.10 &0.35 &0.47 &0.90 &0.63 \\
Ours-TB.-w/o-GCN&2.36&3.11 &0.35 &0.47 &0.91 &0.66 \\
Ours-TB.-GCN&2.19 &\textbf{3.34}&\textbf{0.37}&\textbf{0.50} &\textbf{0.92}&\textbf{0.66} \\
\hline
\end{tabular}
\label{tab2}
\end{table}

From these results, our full model (Ours-TB.-GCN) outperforms others significantly. Two-branch version of our work is superior to the one branch version, because two-branch version fulfills a complementary fusion of details of vision information of RGB frames and semantic context of driving scene. Specially, GCN has the manifest promotion effect for the driver attention prediction because of the semantic context inferring of driving scene. 

Interestingly, SALICON and SalGAN show promising performance. The underlying reason may be two folds. 1) Static methods do not consider the complex motion in prediction, which could reduce the influence of the historical variation of scene that may be considered inappropriately; 2) The behind mechanism of driver attention allocation on temporal videos in challenging situations is complex and still unclear, which can enforce the difficulty to the spatio-temporal attention prediction model. For example, dynamic objects in the background can disturb the prediction easily, as shown by the results of TwoStream in Fig. \ref{fig10}(g). 

In the attention prediction approaches which focus on videos, ACLNet has the closest performance to our method. That is because ACLNet also utilized the Conv-LSTM module to fulfill the temporal attention consideration, while our work output the attention in every 5 frames instead of the one-frame-one-output mode in ACLNet. Therefore, our method can filter the noisy saliency values in prediction, verified by the more focused attention map shown in Fig. \ref{fig10}(c). DR(eye)VE designed a shallow multi-branch 3D encoding module to represent the spatio-temporal variation, which also fused the semantic images to boost the performance. On the contrary, BDDA only with the RGB vision information demonstrates a poor performance. 
\subsection{Evaluation on Driving Accident Clips of DADA}
\begin{table}\vspace{0.3cm}\footnotesize
\centering
\caption{\small{Performance comparison on the testing frames within the accident window (AW) of DADA dataset.}}
\begin{tabular}{ccccccc}
\toprule
Methods/Metrics &      Kldiv &        NSS &        SIM &         CC &      AUC-J &      AUC-S \\
\hline
DR(eye)VE \cite{palazzi2018predicting}  &     3.37 &     1.86 &     0.22 &     0.29 &     0.86 &     0.63 \\
BDDA \cite{xia2017predicting} &     2.50 &     2.63 &     0.29 &     0.40 &     0.89&    {\bf 0.66} \\
TwoStream \cite{bak2017spatio}  &     2.86 &     1.34 &     0.13 &     0.21 &     0.83 &     0.62 \\
MLNet \cite{cornia2016deep} &     11.45 &     0.31 &     0.07 &     0.05 &     0.60 &     0.54 \\
ACLNet \cite{wang2019revisiting}&     2.63 &     2.84 &     0.32 &     0.44 &     0.90 &     0.62 \\
Ours-TB.-GCN &    {\bf2.30} &     {\bf3.12} &     {\bf0.34} &     {\bf0.47} &      {\bf0.91} &     {\bf0.66} \\
\hline
\end{tabular}
\label{tab3}
  \vspace{-0.3cm}
\end{table}

\begin{table*}[htpb]\footnotesize
\centering
\caption{\small{The performance of five dynamic attention prediction methods on three kinds of typical behaviors in accidental scenarios, i.e., the crossing, hitting and out of control of ego vehicle and other crash vehicles. The values in this table are the average of all related videos in the testing set. The number in the bracket is the number of the videos in each behavior type. The best value of each method with respect each metric in different behavior type are marked in \textbf{bold} font.}}
\begin{tabular}{c|c|c|c|c|c|c|c|c|c|c|c|c}
\toprule
\multicolumn{ 1}{c|}{Behavior types} &\multicolumn{ 4}{c|}{\emph{crossing} (42)} & \multicolumn{ 4}{c|}{\emph{hitting} (94)} & \multicolumn{ 4}{c}{\emph{out of control} (19)} \\
\hline
Methods/metrics &     Kldiv &       NSS &       SIM &        CC &     Kldiv&       NSS &       SIM &        CC &     Kldiv &       NSS &       SIM &        CC \\
\hline
 DR(eye)VE \cite{palazzi2018predicting}&    2.43  &    2.69  &    0.29 &    0.41  &    2.28  &    2.78  &    0.31  &    0.43  &    2.09  &    3.00  &    0.31 &    0.45  \\
BDDA \cite{xia2017predicting}&    3.30  &    1.82  &    0.22  &    0.28 &    3.26  &    1.98  &    0.24  &    0.31  &    3.14  &    1.89  &    0.22  &    0.29  \\
MLNet \cite{cornia2016deep}&   11.27  &    0.33  &    0.08  &    0.05  &   12.45  &    0.24  &    0.07  &    0.04  &   11.33  &    0.33  &    0.07  &    0.05  \\
TwoStream \cite{bak2017spatio} &    2.77  &    1.37  &    0.13  &    0.22  &    2.84  &    1.39  &    0.14  &    0.22  &    2.82  &    1.38  &    0.12  &    0.21  \\
ACLNet \cite{wang2019revisiting} &    2.62  &    2.82  &    0.32  &    0.43  &    2.42  &    2.98  &    0.35  &    0.47  &    2.42 &    3.08  &    0.34  &    0.46  \\
Ours-TB.-GCN & {\bf 2.23 } & {\bf 3.16 } & {\bf 0.34 } & {\bf 0.47 } & {\bf 2.10 } & {\bf 3.21 } & {\bf 0.36 } & {\bf 0.49 } & {\bf 1.88 } & {\bf 3.47 } & {\bf 0.37 } & {\bf 0.51 } \\
\hline
\end{tabular}
\label{tab4}
\end{table*}

The testing frames in driving accident window have $14771$ frames. In this evaluation, we compare the proposed method with the attention prediction models that focus on videos, i.e., BDDA \cite{xia2017predicting}, DR(eye)VE \cite{palazzi2018predicting}, TwoStream \cite{bak2017spatio}, MLNet \cite{cornia2016deep}, ACLNet \cite{wang2019revisiting}. Table. \ref{tab3} presents the performance of different methods on all the testing data in accident window of DADA. From the results, we can discover a surprising phenomena that, except from BDDA, the performance of most of methods becomes weaker than that in the overall testing set of DADA.  Our full model also demonstrates a little degradation but still takes the winner.  BDDA focuses on the driver attention prediction in critical situations, which assigns large weight to the frames appeared critical event. Therefore, it shows a good adaptation for the frames in the accident window in DADA.
\begin{figure}[htpb]
\centering
\includegraphics[width=\linewidth]{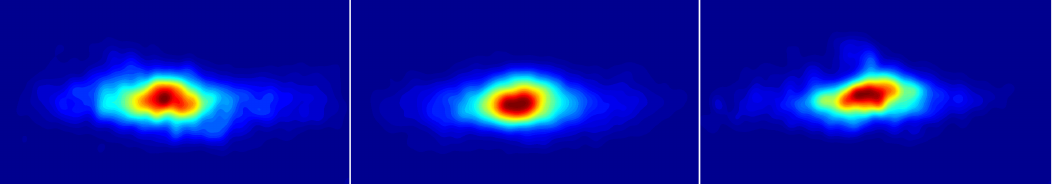}
    \caption{\small{The average attention map of (a) ``\emph{crossing}" behavior, (b) ``\emph{hitting}" behavior and (c) ``\emph{out of control}" in driving accident scenarios in the testing set of DADA dataset.}}
\label{fig11}
  \vspace{-0.1cm}
\end{figure}

 \subsubsection{Comparison \emph{w.r.t.,} Behavior Type in Driving Accidents}
In this work, we also evaluate the performance of video-aware attention prediction methods on different behavior types of driving accidents. In this evaluation, the metrics of KLdiv, NSS, CC, and SIM are adopted. Specifically, we partitioned the video sequences in the testing set into three sets of ``\emph{crossing}", ``\emph{hitting}" and ``\emph{out of control}", where these behavior types are taken by other objects appeared the camera view of ego-car. Table. \ref{tab4} gives the results with respect to each driving behavior type. The results show that our model is the best one over different behavior types. Interestingly, all of the methods demonstrate the worst performance on the ``\emph{crossing}" behavior, and show the best for the ``\emph{out of control}" behavior. That is because, compared with  the ``\emph{hitting}" and ``\emph{out of control}" behaviors,  the objects with ``\emph{crossing}" behavior have apparent scale variation and move slower than that with ``\emph{hitting}" and ``\emph{out of control}". In order to show the behind reason of these phenomena, we demonstrate the average attention map with respect to different behavior types in Fig. \ref{fig11}. From these maps, we can observe that the fixations tend to the middle of the field of vision (FOV) (Fig. \ref{fig11}(b)) for the hitting behavior.  However, crossing behavior is dispersive, and has a longer tail to the sides of FOV (Fig. \ref{fig11}(a)). For the ``out of control" category, the front object often demonstrates a sudden and drastic movement (e.g., the overturned car). Consequently, the largest CC and SIM are obtained by most of methods on the ``\emph{out of control}" behavior.

\begin{figure}[!t]
\centering
\includegraphics[width=\linewidth]{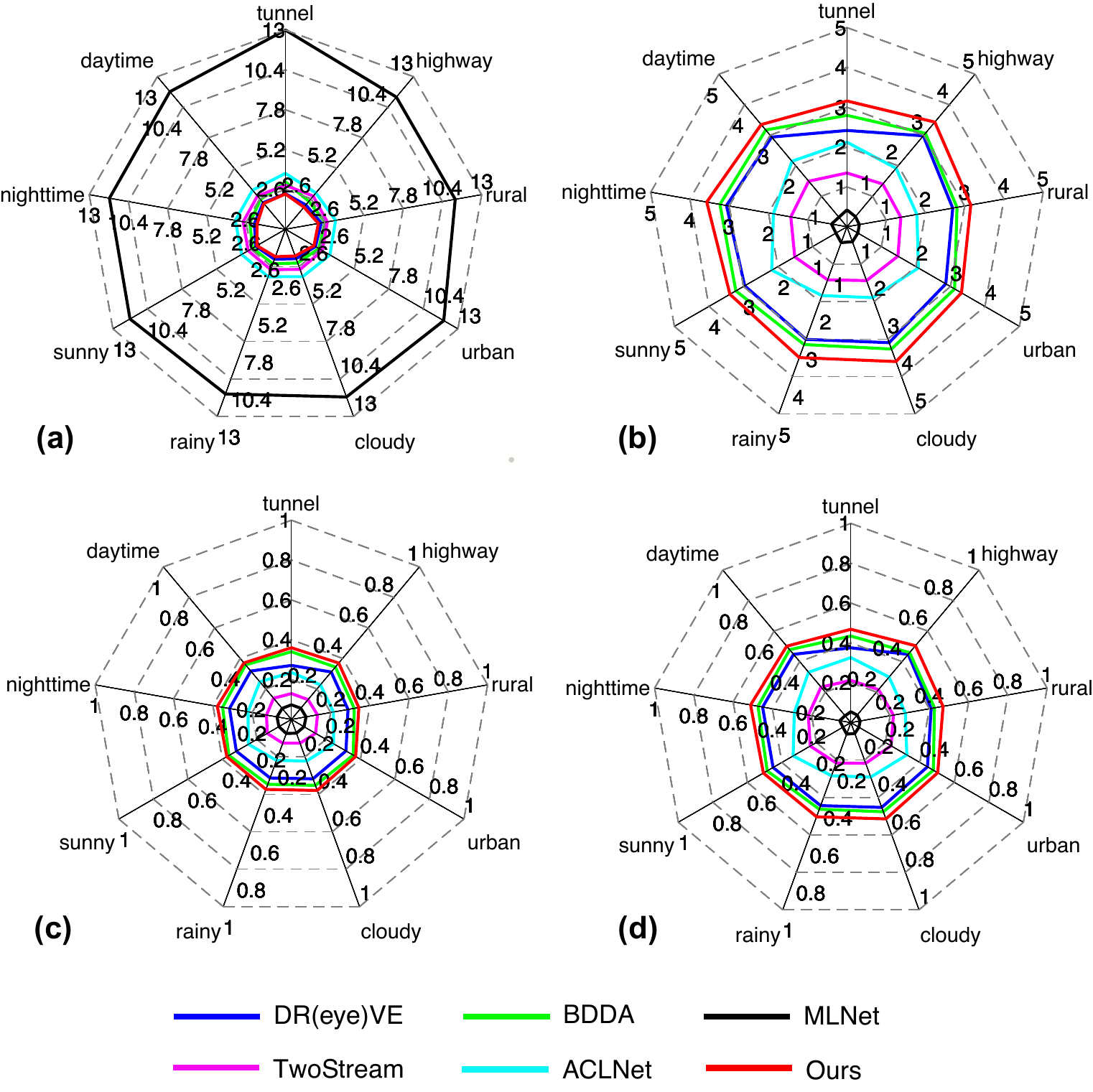}
    \caption{\small{The radar graphs of metrics of (a) KLdiv, (b) NSS, (c) SIM, and (d) CC, w.r.t., distinct methods on various driving situations.}}
\label{fig12}
\end{figure}

\subsubsection{Comparison \emph{w.r.t.,} Different Driving Situations}
For the evaluation of driving attention prediction in driving accident scenarios, we further analyze the performance of different methods on distinct driving situations. In this work, we take the light condition (daytime and nighttime), weather condition (sunny, rainy, and cloudy) and driving occasions (urban, rural, highway, and tunnel) as the situation attributes. Fig. \ref{fig12} presents the radar graphs of different metrics on different methods.
 Fig. \ref{fig12}(a) shows the KLdiv metric, and the inner circles are better than the outer ones. We can see that our method takes the smallest circle, i.e., that our method has stable and superior KLdiv value to others. As for NSS, CC, and SIM metrics, the outer circles are superior to the inner ones, where our method still outperforms the others. BDDA has the closer  performance to ours than other approaches because of the consideration of critical situations.

%\begin{figure}
%\centering
%\includegraphics[width=\linewidth]{fig10.pdf}
%    \caption{\small{The statistics of \# average delayed frames (ADF), w.r.t., 19 kinds of accidental scenarios of humans and our model.}}
%\label{fig13}
%\end{figure}

%\subsubsection{Humans \emph{vs} Our Model for Early Accident Prediction}
%Beside the comparisons on different behaviors and distinct driving situations in driving accident scenarios, we also analyze the capability for early accident prediction by humans and our model. To make a comparable analysis, we compute the average delayed frames (ADF) of humans and our model in the testing set of DADA-2000, and demonstrate the results in Fig. \ref{fig13}. It is worth noting that we have obtained effective results for $19<26$ kinds of driving accident scenarios because the testing set only has 19 kinds of driving accident scenarios with crash-object.  From Fig. \ref{fig13}, we observe that our model has two-side role for different driving accident scenarios, i.e., outperforming humans in some ones and inferior to humans for some other ones. Beside the ``\emph{ego-car hitting motorbike}", the prediction of our model on ego-car involved situations (e.g., ``\emph{ego-car hitting car/truck/pedestrian}") are improved and outperforms humans to a large extent. Additionally, the crossing behaviors of pedestrian, motorbike and cyclist are predicted better than humans, even with an advanced prediction for pedestrian crossing (with a negative ADF).

\subsection{Evaluation on DR(eye)VE and TrafficGaze Datasets.}
In this work, we also evaluate the performance the proposed method on TrafficGaze \cite{deng2019drivers} and DR(eye)VE Datasets \cite{palazzi2018predicting}. TrafficGaze and DR(eye)VE Datasets are all collected from the normal driving situations.  In order to train our model on these two datasets, the semantic images of all the images in DR(eye)VE and TrafficGaze are obtained by Deeplab-V3 semantic segmentation model \cite{DBLP:journals/corr/ChenPSA17}.
\begin{table}\small
\centering
\caption{\small{Performance comparison on TrafficGaze dataset. The best method is labeled by bold font.}}
\begin{tabular}{cccccc}
\toprule
Methods/Metrics &      Kldiv &        NSS &        SIM &         CC &      AUC-J  \\
\hline
MLNet \cite{cornia2016deep} &     5.69 &     0.45 &     0.87 &     0.05 &     0.89 \\
CDNN \cite{deng2019drivers}&     {\bf 0.29} &     5.83&     {\bf 0.77} &    { \bf0.94} &     0.97  \\
Ours-TB.-GCN &    0.66&     {\bf6.10} &     {\bf0.77} &     { \bf0.94} &      {\bf0.98}  \\
\hline
\end{tabular}
\label{tab5}
\end{table}

\begin{table}\small
\centering
\caption{\small{Performance comparison on DR(eye)VE dataset. The best method is labeled by bold font.}}
\begin{tabular}{ccc}
\toprule
Methods/Metrics &      Kldiv &         CC    \\
\hline
MLNet \cite{cornia2016deep} &     2.00 &     0.44 \\
RMDN \cite{DBLP:conf/iclr/BazzaniLT17}&     1.77 &     0.41  \\
DR(eye)VE \cite{palazzi2018predicting} & 1.40 &0.56\\
Ours-TB.-GCN &    {\bf1.35}&     {\bf0.59}  \\
\hline
\end{tabular}
\label{tab6}
\end{table}

For TrafficGaze dataset, we compare the proposed method with MLNet \cite{cornia2016deep} and the newly proposed CDNN \cite{deng2019drivers}. We use the same ground-truth attention maps with the ones adopted in CDNN. In this comparison, the metrics of KLdiv, NSS, CC, SIM, AUC-J are adopted. Table. \ref{tab5} gives the comparisons, where the results of other methods are reported in \cite{deng2019drivers}. From this comparison, we can see that CDNN generates a rather excellent KLdiv value ($0.29$), which means that the predicted attention maps are nearly the same as the ones of ground-truth. Our method shows a better performance than others on other metrics, especially for the NSS ($6.10$). It indicates that the proposed method has a good approximation to the human fixation. Surprisingly, MLNet shows a feasible performance, while it is poor in our DADA dataset. I think the behind meaning is that TrafficGaze is collected from the relatively simple highway scenario while our DADA is complex and full of diverse driving situations.

With respect to DR(eye)VE dataset, it is huge and we test the KLdiv and CC metrics for our method, MLNet \cite{cornia2016deep}, RMDN \cite{DBLP:conf/iclr/BazzaniLT17}, and the full version of DR(eye)VE \cite{palazzi2018predicting}. Table. \ref{tab6} demonstrates the results of each method (The values of other methods are reported in \cite{palazzi2018predicting}), and our method again shows a promising performance. In DR(eye)VE and TrafficGaze datasets, the fixations commonly tend to the vanishing point of the road. Therefore, the lower KLdiv values and higher CC values compared with our DADA are generated.

%\begin{figure}[!t]
%\centering
%\includegraphics[width=\linewidth]{images.pdf}
%    \caption{\small{The predicted attention results by our method on some typical frames of (a) highway occasion (TrafficGaze dataset), (b) urban occasion and (c) nighttime condition (DR(eye)VE dataset). From these results, the fixations tend to the vanishing point of road in TrafficGaze and DR(eye)VE dataset.}}
%\label{fig14}
%\end{figure}

\section{Conclusions}
\label{con}
In this work, we investigated the problem of driver attention prediction in driving accident scenarios. Based on the constructed DADA dataset, we proposed a semantic context induced attentive fusion network (SCAFNet) to fuse the features of RGB frames and semantic context feature of driving scene complementary, and the fused details were transferred over frames by a Conv-LSTM module to find the objects/regions with large motion in driving accident situations. The semantic context feature of driving scene was modeled by a graph convolution network (GCN). Through extensively evaluation on our DADA and other two datasets, i.e., DR(eye)VE and TrafficGaze, the superior performance of SCAFNet to other state-of-the-art methods was obtained. Notably, we analyzed the performance comparison on the testing set within driving accident window from behavior types and driving situations under different light, weather and occasion conditions. The advanced ability of the proposed method was proved in these experiments.
This work achieved a predictive model for driver attention based on the data collected in the laboratory. Based on this work, we want to launch a new question that whether determining driver attention via attention maps can be used within a vehicle to help drivers avoid driving accidents? We will explore its answer in the future.

%References
{\scriptsize{
\bibliographystyle{IEEEtran}
\bibliography{bibfile}}
}

\end{document}